\documentclass{llncs}
\usepackage{pifont}
\usepackage{makeidx}  
\usepackage{graphicx}
\usepackage{color}
\usepackage{amssymb,amsmath,bm}
\usepackage{booktabs}
\usepackage[misc]{ifsym}
\usepackage{multirow}
\usepackage{bbding}
\usepackage{array}
\usepackage{float}
\usepackage{algorithm}
\usepackage{algorithmic}
\newcolumntype{C}[1]{>{\centering\let\newline\\\arraybackslash}m{#1}}

\makeatletter
\newcommand{\printfnsymbol}[1]{%
  \textsuperscript{\@fnsymbol{#1}}%
}
\makeatother
\begin{document}
	\frontmatter          
	%
	%
	\mainmatter              
	\title{A New Bidirectional Unsupervised Domain Adaptation Segmentation Framework}
	
	\author{Munan Ning\thanks{Contributed equally to this work.}\inst{,1,2} \and Cheng Bian\printfnsymbol{1}\textsuperscript{,\Letter}\inst{,1}  \and
    Dong Wei\inst{1} \and Shuang Yu\inst{1} \and Chenglang Yuan\inst{1} \and Yaohua Wang\inst{2} \and Yang Guo\inst{2} \and Kai Ma\inst{1} \and Yefeng Zheng\inst{1}}
    \institute{
    Tencent Jarvis Lab, Shenzhen, China \\
    \and
    National University of Defense Technology, China\\
	\email{tronbian@tencent.com} \\
    }
	
	\maketitle
	\date{}
	\hyphenpenalty=5000
	\tolerance=1000
	
	\begin{abstract}

    Domain shift happens in cross-domain scenarios commonly because of the wide gaps between different domains: when applying a deep learning model well-trained in one domain to another target domain, the model usually performs poorly. 
    To tackle this problem, unsupervised domain adaptation (UDA) techniques are proposed to bridge the gap between different domains, for the purpose of improving model performance without annotation in the target domain.
    Particularly, UDA has a great value for multimodal medical image analysis, where annotation difficulty is a practical concern.
    However, most existing UDA methods can only achieve satisfactory improvements in one adaptation direction (e.g., MRI to CT), but often perform poorly in the other (CT to MRI), limiting their practical usage.
    In this paper, we propose a bidirectional UDA (BiUDA) framework based on disentangled representation learning for equally competent two-way UDA performances.
    This framework employs a unified domain-aware pattern encoder which not only can adaptively encode images in different domains through a domain controller, but also improve model efficiency by eliminating redundant parameters.
    Furthermore, to avoid distortion of contents and patterns of input images during the adaptation process, a content-pattern consistency loss is introduced.
    Additionally, for better UDA segmentation performance, a label consistency strategy is proposed to provide extra supervision by recomposing target-domain-styled images and corresponding source-domain annotations.
    Comparison experiments and ablation studies conducted on two public  datasets demonstrate the superiority of our BiUDA framework to current state-of-the-art UDA methods and the effectiveness of its novel designs.
    By successfully addressing two-way adaptations, our BiUDA framework offers a flexible solution of UDA techniques to the real-world scenario.
	\end{abstract}	
	
\section{Introduction}
Deep learning based models have become dominant for computer-aided automated analysis of medical images and achieved great success in recent years \cite{avendi2016combined,bernard2018deep,fritscher2016deep,kermany2018identifying,litjens2017survey,ronneberger2015u}.
Usually, a great quantity of manual annotations are required for training deep learning models,  yet the annotation process is known to be expertise-demanding, labor-intensive, and time-consuming.
Multimodal imaging---which plays an important role and provides valuable complementary information in disease diagnosis, prognosis, and treatment planning in clinical practice nowadays---may further increase the demand for annotations, as a model well-trained with data of one specific modality often performs poorly on another due to domain shift~\cite{tsai2018learning}.
Unsupervised domain adaptation (UDA)~\cite{ganin2014unsupervised,gong2013connecting} is a quickly rising technique which aims to tackle the domain shift problem by adapting models trained in one domain (the \emph{source} domain) to another (the \emph{target} domain) without manual annotations in the latter~\cite{chang2019all,li2019bidirectional,luo2019taking,tsai2018learning}.
The concept of UDA is readily applicable to the lack-of-annotation problem in multimodal medical imaging~\cite{chen2020unsupervised,dou2018pnp}.
For example, Chen \textit{et al.}~\cite{chen2020unsupervised} applied UDA for model adaptation from cardiac magnetic resonance imaging (MRI) to cardiac computed tomography (CT) based on image-level alignment.
In spite of impressive results obtained for adapting MRI-trained models for CT data~\cite{chen2020unsupervised,dou2018pnp}, few existing methods simultaneously concern the opposite, i.e., CT to MRI.
In practice, situations may arise in which the CT data have already been annotated while the MRI data are not, and the adaptation of CT-trained models for MRI becomes useful.
Therefore, a comprehensive UDA method that is able to effectively accomplish bidirectional adaptations (termed \emph{BiUDA} in this work) is of great clinical value.

Given the importance of BiUDA, we conducted experiments to examine the capabilities of several state-of-the-art (SOTA) UDA methods~\cite{chang2019all,chen2020unsupervised,dou2018pnp,li2019bidirectional,luo2019taking,tsai2018learning} for bidirectional adaptations between MRI and CT using the Multi-Modality Whole Heart Segmentation (MMWHS) challenge dataset~\cite{zhuang2016multi} and the Multi-Modality Abdominal Segmentation (MMAS) dataset~\cite{kavur2020chaos,landman2015multi}.
Taking MMWHS dataset as an example, we show the bidirectional adaptation results in  Fig.~\ref{fig:Histogram}. In general, these methods worked effectively for adapting MRI-trained models for CT data. When reversing the adaptation direction, however, most of them suffered a dramatic drop in performance and failed to produce comparable results. Similar results can also be found on the MMAS dataset.
We call this phenomenon \emph{domain drop}, which intuitively reflects the substantial performance gap between the bidirectional adaptations in BiUDA.
Recently, Dou \textit{et al.}~\cite{dou2018pnp} improved performance of the CT-MRI BiUDA by switching the early layers of the two encoders to fit individual modality and sharing the higher layers between two modalities.
It is presumed that the improvement was due to the isolation of low-level features of each modality.
Motivated by the presumption, we propose a novel framework based on the concept of disentangled representation learning (DRPL)~\cite{huang2018multimodal,lee2018diverse}, to address the domain drop problem in BiUDA.
Specifically, our framework decomposes an input image into a content code and a pattern code, with the former representing domain-invariant characteristics shared by different modalities (e.g., shape) and the latter representing domain-specific features isolated from each other (e.g., appearance).

Our main contributions include:
\begin{itemize}
\item[1)] We propose a novel BiUDA framework based on DRPL, which effectively mitigates the domain drop problem and achieves SOTA results for both adaptation directions in experiments on the MMWHS and MMAS datasets.
\item[2)] Unlike existing works~\cite{chang2019all,huang2018multimodal,lee2018diverse} that adopted a separate pattern encoder for each modality, we design a
domain-aware pattern encoder 
to unify the encoding process of different domains in a single module with a domain controller.
This design not only reduces parameters but also improves performance of the network.
In addition, a content-pattern consistency loss is proposed to avoid the distortion of the content and pattern codes.
\item[3)] For better performance in the UDA segmentation task, we propose a label consistency loss, utilizing the target-domain-styled images
    and corresponding source-domain annotations for auxiliary training.
    \end{itemize}

\section{Method}

\subsection{Problem Definition}
In the UDA segmentation problem,
there is an annotated dataset $\left\{\left(x^{i}_{s}, a^{i}_{s}\right)\right\}_{i=1}^{N_{s}}$ in the source domain $\mathcal{X}_{s}$, where each image $x_s^i$ has a unique pixel-wise annotation $a_s^i$.
Meanwhile, there exists an unannotated dataset $\left\{x^{i}_{t}\right\}_{i=1}^{N_{t}}$ in the target domain $\mathcal{X}_{t}$.
The goal of our framework is to utilize the annotated source-domain data to obtain a model that is able to perform well on the unannotated target-domain data.
We shall omit the superscript index $i$ and superscript domain indicators $s$ and $t$ for simplicity in case of no confusion.

\begin{figure}[ht]
	\centering
	\includegraphics[width=0.9\columnwidth]{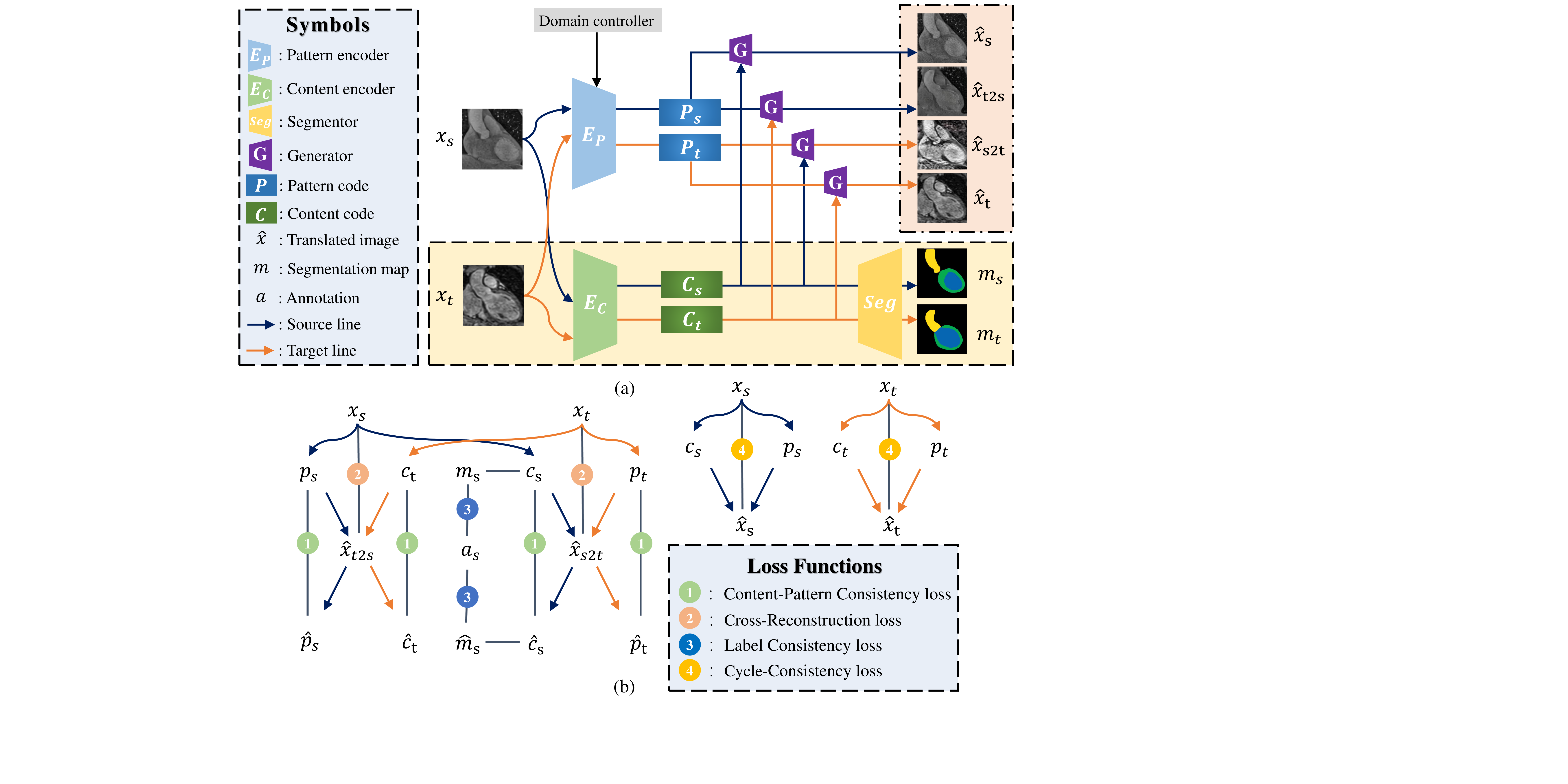}
	\caption{ (a) Framework diagram;  (b) Data flow and corresponding loss functions.}
	\label{fig:Framework}
    \end{figure}

Fig.~\ref{fig:Framework}(a) shows the diagram of our DRPL-based BiUDA framework equipped with the domain-aware pattern encoder, which will be elaborated in Section~\ref{sec:DRPL}.
Fig.~\ref{fig:Framework}(b) illustrates the information flow and corresponding loss computations within the proposed framework, which will be elaborated in Section~\ref{sec:losses}.

\subsection{DRPL Framework with Domain-Aware Pattern Encoder}
\label{sec:DRPL}
Compared to the existing UDA methods that attempted to align the features extracted in different domains, we argue that a better solution is to explicitly make the model aware of the cross-domain commonalities and differences.
For this reason, we adopt the DRPL~\cite{huang2018multimodal,lee2018diverse} framework to disentangle the image space into a domain-sharing content space $\mathcal{C}$ (implying anatomical structures) and a domain-specific pattern space $\mathcal{P}$ (implying appearances), using a content encoder $E_c$ and a pattern encoder $E_p$ (see Fig.~\ref{fig:Framework}(a)), respectively.
Concretely, $E_{c}: \mathcal{X} \rightarrow \mathcal{C}$ maps an input image to its content code: $\boldsymbol{c}=E_c(x)$, and $E_{p}: \mathcal{X} \rightarrow \mathcal{P}$ maps the image to its pattern code.
It is worth mentioning that as the pattern codes are domain-specific, a common strategy is to employ dual pattern encoders---one for each domain~\cite{chang2019all,huang2018multimodal,lee2018diverse}.
On the contrary, we propose a unified domain-aware pattern encoder for both domains, which is controlled by a domain controller $d\in\{0,1\}$, where 0 and 1 indicate the source and target domains, respectively.
Hence, by specifying $d$, $E_p$ adaptively encodes images from different domains into representative pattern codes: $\boldsymbol{p}_s=E_p(x|0)$ and $\boldsymbol{p}_t=E_p(x|1)$.
The proposed $E_p(x|d)$ simplifies our network design, greatly reducing the number of parameters to learn during training. Moreover, the proposed domain controller improves the encoding ability since it forces the pattern encoder to learn the differences between two domains by providing additional pattern information.

After being extracted from the input images, the content and pattern codes in different domains are permuted and recombined, and the resulting pairs are input to a generator $G:(\mathcal{C}, \mathcal{P}) \rightarrow \hat{\mathcal{X}}$ to recompose images with the anatomical structures specified by \boldsymbol{$c$} and appearance specified by \boldsymbol{$p$}: $\hat{x}=G(\boldsymbol{c},\boldsymbol{p})$.
Here, $\hat{\cdot}$ indicates the variable is within or resulted from the recomposed image space.
Four types of images can be recomposed based on the permutation of \boldsymbol{$c$} and \boldsymbol{$p$}, i.e., reconstructed source image $\hat{x}_s$ (both \boldsymbol{$c$} and \boldsymbol{$p$} from the source domain), reconstructed target image $\hat{x}_t$ (both \boldsymbol{$c$} and \boldsymbol{$p$} from the target domain),
and translated images $\hat{x}_{s2t}$ (\boldsymbol{$c$} from the source and \boldsymbol{$p$} from the target domain) and $\hat{x}_{t2s}$ (\boldsymbol{$c$} from the target and \boldsymbol{$p$} from the source domain);
here, `s2t' stands for `source to target', and vice versa.
Note that a single generator is used for the recomposition of the four types of images.
See the generator and recomposed images in Fig.~\ref{fig:Framework}(a).
Lastly, a segmenter $S$ is employed to decode the content codes \boldsymbol{$c$} to semantic segmentation masks ${m}$: $m = S\left(\boldsymbol{c}\right)$.

\subsection{Loss Functions for DRPL-based BiUDA Framework}
\label{sec:losses}
\textbf{Content-pattern consistency loss.}
When performing domain transfer with many existing DRPL methods (e.g.,~\cite{huang2018multimodal} and~\cite{lee2018diverse}), we notice apparent anatomical and/or texture distortions in the recomposed images.
We assume the fundamental reason to be the distortion of the content and pattern codes while going through the decomposition-recomposition process.
To avoid such distortion, we propose a content-pattern consistency loss to penalize potential distortions of \boldsymbol{$c$} and \boldsymbol{$p$} in the workflow.
Let $\boldsymbol{\hat{c}}=E_{c}\left(\hat{x}\right)$ and $\boldsymbol{\hat{p}}=E_{p}\left(\hat{x}, d\right)$ denote the content and pattern codes for the \emph{reconstructed} image $\hat{x}$ (i.e., $\hat{x}_s$ or $\hat{x}_t$), and $\boldsymbol{c}=E_{c}\left(x\right)$ and $\boldsymbol{p}=E_{p}\left(x,d\right)$ for the input image $x$.
Then, the content-pattern consistency loss $\mathcal{L}^{cpc}$ (represented by \ding{172} in Fig.~\ref{fig:Framework}(b)) is formulated as:
\begin{equation}
 \mathcal{L}^{cpc}=\mathbb{E}_{\boldsymbol{\hat{c}} \sim \hat{\mathcal{C}}, \boldsymbol{c} \sim \mathcal{C}}\left[\left\|\boldsymbol{\hat{c}}-\boldsymbol{c}\right\|_{1}\right]
 +\mathbb{E}_{\boldsymbol{\hat{p}} \sim \hat{\mathcal{P}}, \boldsymbol{p} \sim \mathcal{P}}\left[\left\|\boldsymbol{\hat{p}}-\boldsymbol{p}\right\|_{1}\right],
\end{equation}
where $\|\cdot\|_1$ represents the L1 norm.

\textbf{Label consistency loss.}
In our proposed DRPL framework, ideally, the source-domain image $x_s$ and source-to-target transferred image $\hat{x}_{s2t}$ should contain the same anatomical structures, since the latter is recomposed with the content code of the former.
In addition, their anatomical consistency is further enhanced by the content consistency loss described above.
Therefore, a label consistency loss is introduced to supervise the segmentation of both $x_s$ and $\hat{x}_{s2t}$ with the same annotation $a_{s}$.
Let $m_{s}=S(c_s)$ and $\hat{m}_s=S(\hat{c}_s)$ denote the segmentation masks of $x_s$ and $\hat{x}_{s2t}$, respectively, where $c_{s}=E_c(x_s)$ and $\hat{c}_{s}=E_c(\hat{x}_{s2t})$.
Then, the segmentation masks can be supervised by $a_{s}$ using a combination of the cross-entropy and Dice losses:
\begin{equation}
    \mathcal{L}^{seg}\left(a_s, m\right)=1
    -\frac{1}{N} \sum_{j} a_s(j) \log m(j)
    -\sum_{j} \frac{2  a_s(j) m(j)}{ a^{2}_s(j)+m^{2}(j)},
\end{equation}
where $j$ iterates over all locations and channels in $a_s$ and $m$, and $N$ is the total number of iterations.
Accordingly, the proposed label consistency loss (represented by \ding{174} in Fig.~\ref{fig:Framework}(b)) is defined as:
\begin{equation}\label{eq:label_consist}
 \mathcal{L}^{lc}=\mathcal{L}^{seg}(a_s,m_s)+\mathcal{L}^{seg}(a_s,\hat{m}_s).
\end{equation}
It is worth noting that $\mathcal{L}^{seg}(a_s,\hat{m}_s)$ in Eq. (\ref{eq:label_consist}) can be viewed as providing supplementary target-domain training data to make up the vacancy of annotation in the target domain.
Therefore, it is expected to help relieve the domain drop.

\textbf{Cycle-consistency and cross-reconstruction losses.}
Following the intuition in CycleGAN~\cite{zhu2017unpaired}, the input images and their reconstructions are constrained to be close with a cycle-consistency loss (represented by \ding{175} in Fig.~\ref{fig:Framework}(b)):
\begin{equation}\label{eq:l_cycle}
\begin{split}
    \mathcal{L}^{cycle}=\mathbb{E}_{\hat{x} \sim \hat{\mathcal{X}}, x \sim \mathcal{X}}\left[\left\|\hat{x}-x\right\|_{1}\right].
\end{split}
\end{equation}
Note that $\hat{x}$ in Eq.~(\ref{eq:l_cycle}) should only be reconstructed images, i.e., either $\hat{x}_s$ or $\hat{x}_t$.
In addition, the translated images $\hat{x}_{s2t}$ should be indistinguishable from the real target-domain images $x_t$, to provide $\mathcal{L}^{seg}(a_s,\hat{m}_s)$ with high-quality recomposed target-domain data.
Following the generative adversarial network (GAN) ~\cite{goodfellow2014generative}, a discriminator $D$ is introduced and the cross-reconstruction loss (represented by \ding{173} in Fig.~\ref{fig:Framework}(b)) is defined as:
\begin{equation}
\begin{split}
    \mathcal{L}^{GAN}_{\hat{x}_{s2t}}=\mathbb{E}_{\hat{x}_{s2t} \sim \hat{\mathcal{X}}_{s2t}}\left[\log \left(1-D\left(\hat{x}_{s2t}\right)\right)\right]+\mathbb{E}_{x_{t} \sim \mathcal{X}_{t}}\left[\log D\left(x_{t}\right)\right].
\end{split}
\end{equation}
Likewise, a cross-reconstruction loss $\mathcal{L}^{GAN}_{\hat{x}_{t2s}}$ is also proposed for the target-to-source transferred images $\hat{x}_{t2s}$ as:
\begin{equation}
\begin{split}
    \mathcal{L}^{GAN}_{\hat{x}_{t2s}}=\mathbb{E}_{\hat{x}_{t2s} \sim \hat{\mathcal{X}}_{t2s}}\left[\log \left(1-D\left(\hat{x}_{t2s}\right)\right)\right]+\mathbb{E}_{x_{s} \sim \mathcal{X}_{s}}\left[\log D\left(x_{s}\right)\right].
\end{split}
\end{equation}

\textbf{Overall loss function.}
The overall loss function of our BiUDA framework is a weighted summation of the above-described losses ($\mathcal{L}^{cpc}$ and $\mathcal{L}^{cycle}$ are computed in both the source and target domains):
    \begin{equation}
    \label{eq:lambda}
    \mathcal{L}=\lambda_1 (\mathcal{L}^{cpc}_{s}+\mathcal{L}^{cpc}_{t})
    +\lambda_2 \mathcal{L}^{lc}
    +\lambda_3 (\mathcal{L}^{cycle}_{s}+\mathcal{L}^{cycle}_{t})
    +\lambda_4 (\mathcal{L}^{GAN}_{\hat{x}_{s2t}}+\mathcal{L}^{GAN}_{\hat{x}_{t2s}}).
    \end{equation}
    
\section{Experiments}
\textbf{Datasets.}
The proposed BiUDA framework is evaluated using the MMWHS challenge dataset~\cite{zhuang2016multi} and the MMAS dataset.
The MMWHS dataset includes 20 MRI (47--127 slices per scan) and 20 CT (142--251 slices per scan) cardiac scans.
Four cardiac anatomic structures, including ascending aorta (AA), left atrium blood cavity (LABC), left ventricle blood cavity (LVBC), and left ventricle myocardium (LVM), are annotated.
The MMAS dataset includes 20 MRI scans (21--33 slices per scan) from the CHAOS Challenge~\cite{kavur2020chaos} and 30 CT scans (35--117 slices per scan) from \cite{landman2015multi}. Multiple organs are manually annotated, including liver, right kidney (R-Kid), left kidney (L-Kid), and spleen. For a fair comparison, every input slice is resized to 256$\times$256 pixels and augmented in the same way as SIFA~\cite{chen2020unsupervised}, including random crop, flip and rotation. A 5-fold cross-validation strategy is employed to test our framework.

\textbf{Evaluation metrics.}
The Dice coefficient (Dice) and F1 score are used as the basic evaluation metrics.
The performance upper-bounds are established by separately training and testing two segmentation networks (one for each modality) using data of the same modality, and denoted by `M2M' (MRI to MRI) and `C2C' (CT to CT), respectively.
Followed by the same adaptation direction reported in ~\cite{chen2020unsupervised}, we define the adaptation from MRI (as suorce domain) to CT (as target domain) as the forward adaptation, and vice versa.
Since the levels of difficulty are markedly different for MRI- and CT-based cardiac segmentation due to distinct modal characteristics (see the rows for M2M and C2C in Table~\ref{table:quanti_metric}), it would be difficult to directly compare Dice or F1 scores obtained via the forward and backward adaptations.
Instead, we resort to the performance drop, which is computed by subtracting the UDA performance from the corresponding target domain upper-bound, e.g., subtracting MRI-to-CT UDA performance from C2C.
Lastly, to intuitively reflect the quality of BiUDA using a single metric, we further calculate the average performance drop by averaging the bidirectional performance drops.
	\begin{table*}[t]
    \caption{
        Performance comparison of our proposed BiUDA framework with SOTA UDA algorithms on the MMWHS dataset using the average performance drops (lower is better) in Dice and F1 score.
         \label{table:quanti_metric}}
    	\centering
    	\renewcommand{\arraystretch}{1.0}
    	\scalebox{0.65}{
    	\large
    	\begin{tabular}{c|C{1.4cm}|C{1.4cm}|C{1.4cm}|C{1.4cm}|C{1.4cm}|C{1.4cm}|C{1.4cm}|C{1.4cm}|C{1.4cm}|C{1.4cm}}			
    		\toprule[2pt]
    		\multirow{2}{*}{\bf{Method}} & \multicolumn{2}{C{2.8cm}|}{\bf{AA}}& \multicolumn{2}{C{2.8cm}|}{\bf{LABC}}& \multicolumn{2}{C{2.8cm}|}{\bf{LVBC}}& \multicolumn{2}{C{2.8cm}|}{\bf{LVM}}& \multicolumn{2}{C{2.8cm}}{\bf{Mean}}\\
    		\cline{2-11}		
    		&\textbf{Dice$^{\downarrow}$} 	&\textbf{F1$^{\downarrow}$} 	 	&\textbf{Dice$^{\downarrow}$}   &\textbf{F1$^{\downarrow}$}	  &\textbf{Dice$^{\downarrow}$}   &\textbf{F1$^{\downarrow}$} 	  &\textbf{Dice$^{\downarrow}$}   &\textbf{F1$^{\downarrow}$} 	
    		&\textbf{Dice$^{\downarrow}$}   &\textbf{F1$^{\downarrow}$}  \\
    		\cline{0-1}
    		\cline{2-11}
    		M2M$^{*}$  &81.68 &82.11   &85.25 &85.39   &93.01 &93.07  &84.68 &84.71  &86.16  &86.39\\
            C2C$^{*}$  &96.17 &96.21   &93.25 &93.28    &89.67  &89.92   &84.26  &84.54   &90.84  &91.07\\
    		\hline
    		
    		AdaptSegNet~\cite{tsai2018learning}  &55.80   &52.50    &48.94  &48.67  &23.20  &19.62     &34.14  &33.41     &40.52  &37.62  \\
    		BDL ~\cite{li2019bidirectional} &55.31 	&48.06 	&48.91 	&47.88 	&32.33 	&25.42 	&44.47 	&44.15 	&45.25 	&40.15   \\
    		CLAN~\cite{luo2019taking}  &56.90 	&51.46 	&49.17 	&47.47 	&23.18 	&19.58 	&33.49 	&32.76 	&40.68 	&36.68   \\
    		DISE~\cite{chang2019all} &38.48 	&34.89 	&43.70 	&34.39 	&12.61 	&10.66 	&32.62 	&27.31 	&31.85 	&25.19   \\
            SIFA~\cite{chen2020unsupervised} &16.29 &15.02 &22.72 &21.58 &20.61 &19.42 &27.74 &27.05 &21.84 &20.49   \\
            ACE~\cite{wu2019ace} &13.39 &11.48 &11.43 &10.81 &5.08 &4.76 &24.00 &23.31 &13.47 &12.44  \\
    		Ours &\textbf{8.70} 	&\textbf{8.54} 	&\textbf{6.68} 	&\textbf{6.37} 	&\textbf{3.50} 	&\textbf{3.48} 	&\textbf{15.07} 	&\textbf{14.76} 	&\textbf{8.49} 	&\textbf{8.29}   \\
    		\bottomrule[2pt]
    		\multicolumn{11}{l}{$^{*}$\footnotesize{Upper-bound performances are reported as the original Dice and F1 scores.}} \\
    	\end{tabular}}
    \end{table*}

	\begin{table*}[t]
    \caption{
        Performance comparison of our proposed BiUDA framework with SOTA UDA algorithms on the MMAS dataset.
        Dice$^{\downarrow}$: average performance drop in Dice; F1$^{\downarrow}$: average performance drop in F1 score. 
         \label{table:quanti_metric1}}
    	\centering
    	\renewcommand{\arraystretch}{1.0}
    	\scalebox{0.65}{
    	\large
    	\begin{tabular}{c|C{1.4cm}|C{1.4cm}|C{1.4cm}|C{1.4cm}|C{1.4cm}|C{1.4cm}|C{1.4cm}|C{1.4cm}|C{1.4cm}|C{1.4cm}}			
    		\toprule[2pt]
    		\multirow{2}{*}{\bf{Method}} & \multicolumn{2}{C{2.8cm}|}{\bf{Liver}}& \multicolumn{2}{C{2.8cm}|}{\bf{R.kidney}}& \multicolumn{2}{C{2.8cm}|}{\bf{L.kidney}}& \multicolumn{2}{C{2.8cm}|}{\bf{Spleen}}& \multicolumn{2}{C{2.8cm}}{\bf{Mean}}\\
    		\cline{2-11}		
    		&\textbf{Dice$^{\downarrow}$} 	&\textbf{F1$^{\downarrow}$} 	 	&\textbf{Dice$^{\downarrow}$}   &\textbf{F1$^{\downarrow}$}	  &\textbf{Dice$^{\downarrow}$}   &\textbf{F1$^{\downarrow}$} 	  &\textbf{Dice$^{\downarrow}$}   &\textbf{F1$^{\downarrow}$} 	
    		&\textbf{Dice$^{\downarrow}$}   &\textbf{F1$^{\downarrow}$}  \\
    		\cline{0-1}
    		\cline{2-11}
    		M2M$^{*}$  &93.89 	&93.98	&93.34 	&93.16	&92.30 	&92.13	&91.95 	&92.6	&92.87 	&92.97 
\\
            C2C$^{*}$  &96.25 	&96.06	&90.99 	&90.68	&91.86 	&92.27	&93.72 	&92.96	&93.21 	&92.99 
\\
    		\hline
    		
    		AdaptSegNet~\cite{tsai2018learning}  &14.29 	&17.87 	&27.33 	&32.58 	&39.52 	&43.36 	&27.36 	&32.46 	&27.12 	&31.57 \\
    		BDL ~\cite{li2019bidirectional} &21.94 	&26.81 	&27.93 	&32.59 	&44.18 	&48.79 	&27.29 	&30.91 	&30.33 	&34.77 \\
    		CLAN~\cite{luo2019taking} &19.90 	&24.80 	&27.63 	&32.61 	&40.37 	&44.07 	&27.37 	&32.57 	&28.82 	&33.51 \\
    		DISE~\cite{chang2019all} &8.00 	&10.04 	&9.38 	&11.24 	&10.01 	&12.34 	&8.27 	&10.71 	&8.92 	&11.08 \\
            SIFA~\cite{chen2020unsupervised} &5.84 	&5.93 	&5.75 	&5.93 	&9.94 	&10.94 	&8.55 	&8.96 	&7.52 	&7.94 \\
            ACE~\cite{wu2019ace} &5.36 	&5.90 	&5.82 	&7.16 	&5.02 	&5.99 	&4.89 	&5.57 	&5.27 	&6.15 \\
    		Ours &\textbf{5.04} 	&\textbf{5.79} 	&\textbf{4.44} 	&\textbf{5.13} 	&\textbf{3.69} 	&\textbf{4.42} 	&\textbf{4.38} 	&\textbf{5.42} 	&\textbf{4.39} 	&\textbf{5.19}   \\
    		\bottomrule[2pt]
    		\multicolumn{11}{l}{$^{*}$\footnotesize{Upper-bound performances are reported as the original Dice and F1 scores.}} \\
    	\end{tabular}}
    \end{table*}

\textbf{Implementation.}
The content encoder $E_{c}$ is based on PSP-101~\cite{zhao2017pyramid}, accompanied by a fully convolutional network as the segmenter $S$.
The domain-aware pattern encoder $E_{p}$ comprises several downsampling units followed by a global average pooling layer and a $1 \times 1$ convolution layer.
Each unit contains a convolution layer with $stride = 2$, followed by a batch normalization layer and a ReLU layer.
The discriminator $D$ has a similar structure to $E_{p}$, with its units consisting of a convolution layer with $stride = 2$, an instance normalization layer and a leaky-ReLU layer.
The generator $G$ is composed of a set of residual blocks with adaptive instance normalization (AdaIN)~\cite{huang2017arbitrary} layers and several upsampling and convolution layers.
During the inference stage, only $E_{c}$ and $S$ are used to obtain the segmentation results. The whole framework is implemented with PyTorch on an NVIDIA Tesla P40 GPU.
We use a mini-batch size of 8 for training, and train the framework for 30,000 iterations.
We use the SGD optimizer with an initial learning rate of $2.5 \times 10^{-4}$ for $E_{c}$, and the Adam optimizer with an initial learning rate of $1.0 \times 10^{-3}$ for $E_{p}$ and $G$.
The alternating training scheme~\cite{goodfellow2014generative} is adopted to train the discriminator $D$ using the Adam optimizer with an initial learning rate of $1.0 \times 10^{-4}$.
The polynomial decay policy is adopted to adjust all learning rates.
The hyper-parameters $\lambda_1, \lambda_2, \lambda_3$ and $\lambda_4$ in Eq. (\ref{eq:lambda}) 
are empirically set to 0.01, 1.0, 0.5, and 0.01, respectively, although we find in our experiments that the results are not very sensitive with respect to the exact values of these parameters.
The upper-bound networks for M2M and C2C are implemented using the standard PSP101 backbone and trained with the same settings as $E_{c}$.

	\begin{figure}[t]
	\centering
	\includegraphics[width=0.9\columnwidth]{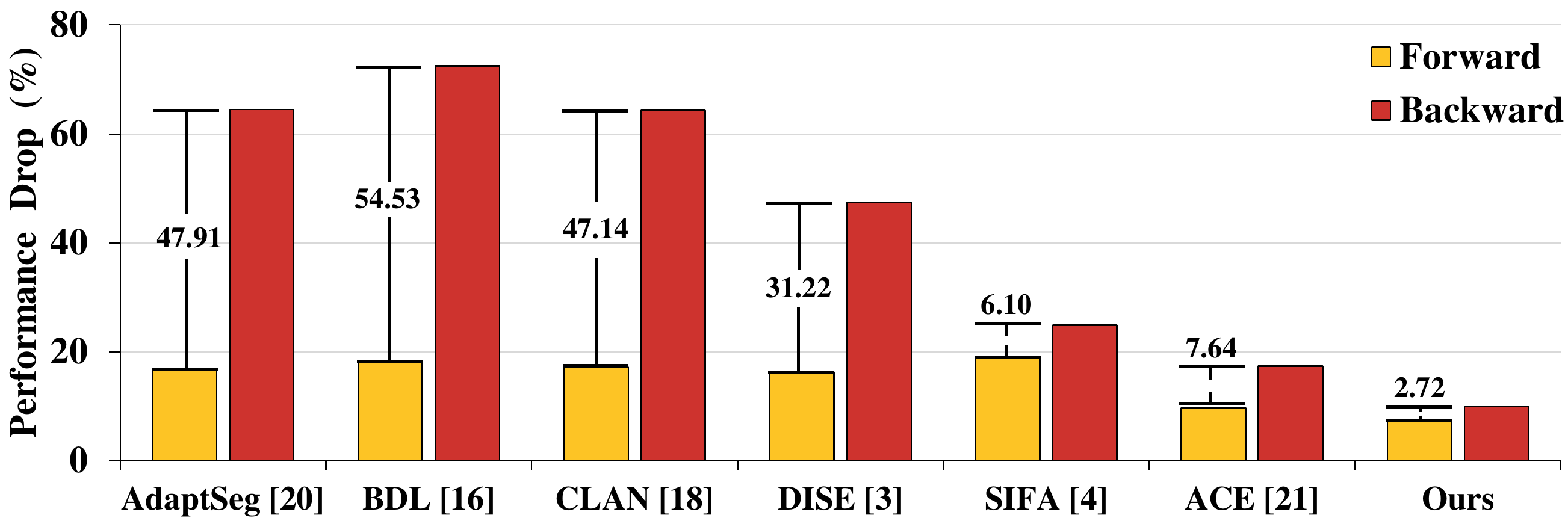}
	\caption{MMWHS: the visualizaiton of the performance drops in Dice in the forward (yellow bars) and backward (red bars) adaptation directions (lower is better), and the gap between the bidirectional performance drops for each method (narrower is better) on the MMWHS dataset. It proves that our method achieved SOTA results for both adaptation directions and comparable performance gap.}
	\label{fig:Histogram}
    \end{figure}
	
\begin{figure}[t]
\centering
\includegraphics[width=0.9\columnwidth]{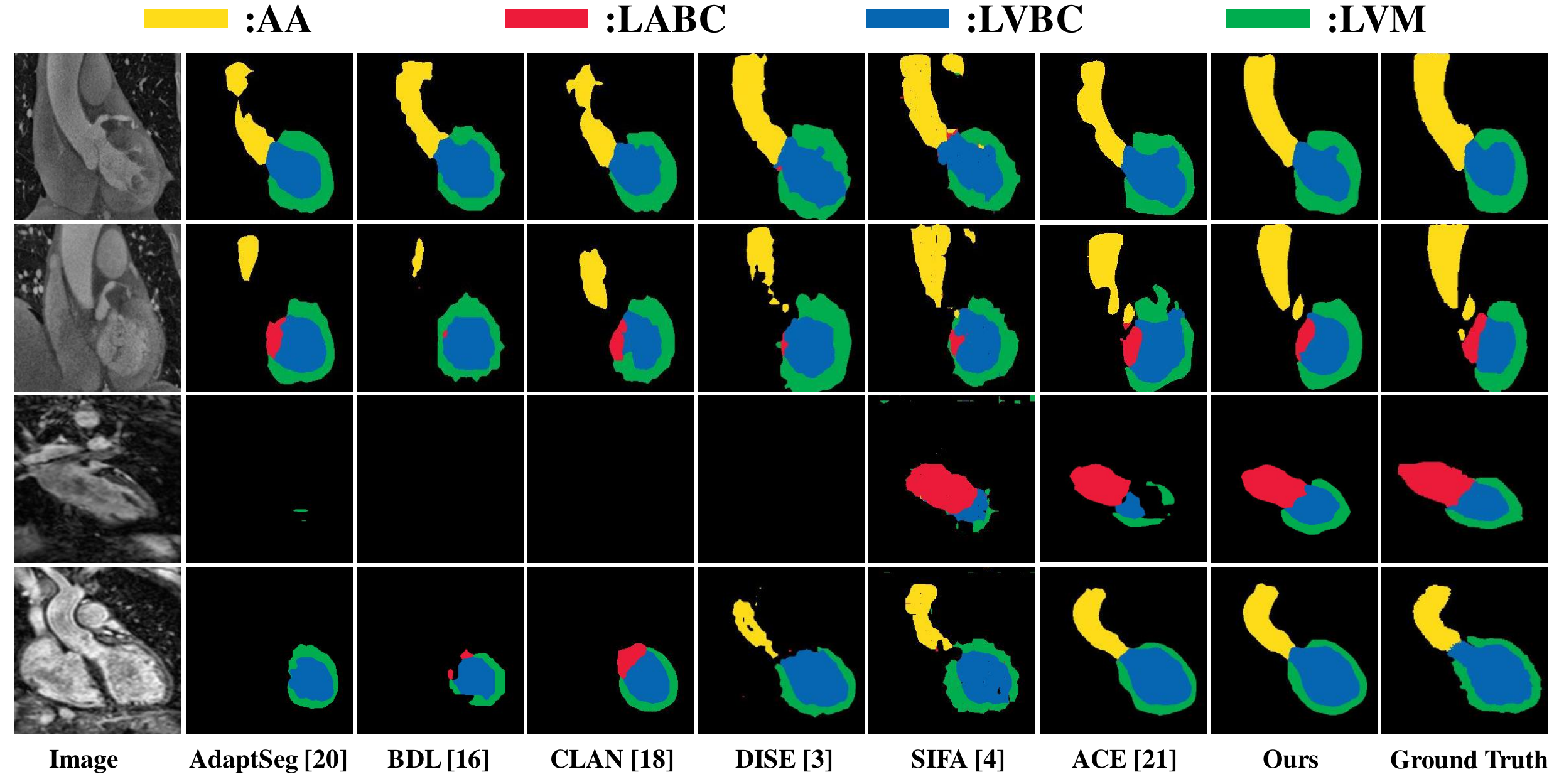}
\caption{Illustration of the BiUDA segmentation results by different methods on the MMWHS dataset~\cite{zhuang2016multi}.
The top two rows show the forward adaptation, while the bottom rows are the backward adaptations, respectively.}
\label{fig:Visualization_MMWHS}
\end{figure}

\begin{figure}[t]
\centering
\includegraphics[width=0.9\columnwidth]{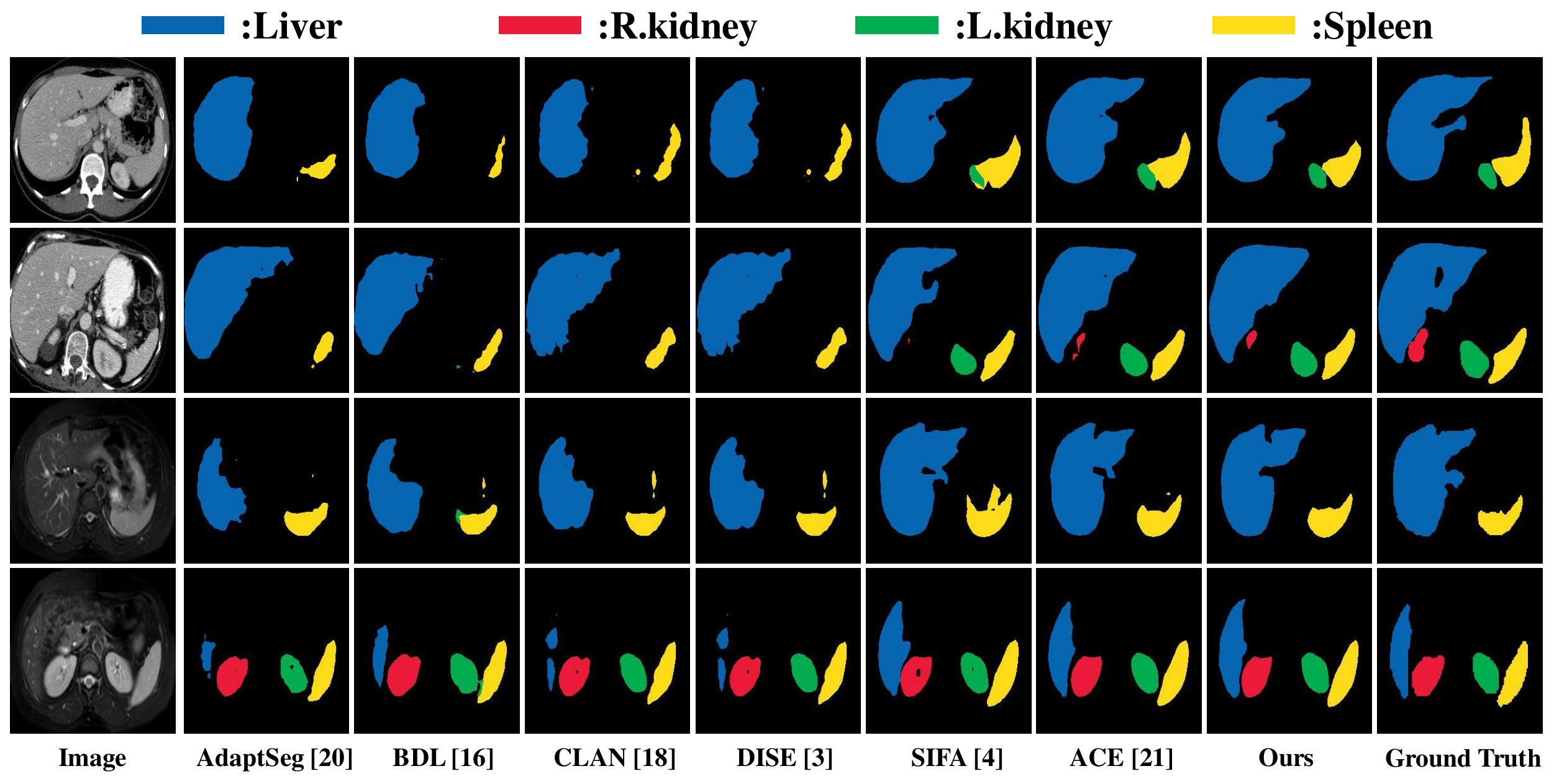}
\caption{Illustration of the BiUDA segmentation results by different methods on the MMAS dataset~\cite{kavur2020chaos,landman2015multi}.
The top two rows show the forward adaptation, while the bottom rows are the backward adaptations, respectively.}
\label{fig:Visualization_CHAOS}
\end{figure}

\textbf{Quantitative and qualitative analyses.}
To validate the efficacy of our framework in addressing the domain drop problem, extensive experiments are conducted on the two datasets.
The competing algorithms include several SOTA UDA methods ~\cite{chang2019all,chen2020unsupervised,dou2018pnp,li2019bidirectional,luo2019taking,tsai2018learning} in both the computer vision and medical image fields.
We reimplement all compared methods with the origin released codes and apply with the default configurations. 
Table~\ref{table:quanti_metric} presents the average performance drops in Dice and F1 score for the four cardiac structures as well as the mean values across structures on the MMWHS dataset. Table~\ref{table:quanti_metric1} presents the average performance drops for four abdominal structures and the mean values across them on the MMAS dataset.
As can be seen, our framework outperforms all competing methods by large margins for all structures on these datasets.
In addition, Fig.~\ref{fig:Histogram} visualizes the forward and backward performance drops and corresponding gaps between the bidirectional drops on the MMWHS dataset.
For most competing UDA methods, there exist considerable gaps between the performances of the forward and backward adaptations.
In contrast, our method significantly narrows this gap, and achieves the lowest performance drops in both adaptation directions.
To further show the effectiveness of our method, we visualize the segmentation results by ours and competing methods in Fig.~\ref{fig:Visualization_MMWHS} and Fig.~\ref{fig:Visualization_CHAOS} for qualitative comparisons.
As we can see, the segmentations by our method are much closer to the ground truth, especially for the backward adaptation.
To summarise, the comparative experiments indicate that our framework can effectively address the domain drop problem, which has been overlooked by other UDA methods.
Accordingly, our proposed framework presents new SOTA results for BiUDA segmentation on both the MMWHS and MMAS datasets.

\textbf{Ablation study.}
We conduct ablation studies with incremental evaluations to validate the effectiveness of our novel modules, 
including the content-pattern consistency loss, the label consistency loss, and the domain-aware pattern encoder.
The results on the MMWHS dataset are shown in Table~\ref{table:abstudy}.
As we can see, after adding the content-pattern consistency loss, the average performance drop in Dice is reduced to 25.73\%.
In addition, the introduction of the label consistency loss further reduces the average drop to 12.98\%, benefiting from pseudo training pairs in the target domain.
Lastly, with the domain controller, the unified domain-aware pattern encoder learns pattern information from different modalities simultaneously, further reducing the average drop to 8.49\%.

    \begin{table}[t]
	\centering
	\caption{Ablation studies of our proposed modules on the MMWHS dataset.
The bidirectional average performance drop in Dice is used for evaluation.
}\label{table:abstudy}
	\normalsize
	\renewcommand{\arraystretch}{1.1}
	\scalebox{0.65}{
	\large
	\begin{tabular}{p{3.5cm}<{\centering}|p{1.3cm}<{\centering}p{1.3cm}<{\centering}p{1.3cm}<{\centering}p{1.3cm}<{\centering}|p{1.5cm}<{\centering}p{1.5cm}<{\centering}p{1.5cm}<{\centering}p{1.5cm}<{\centering}p{1.5cm}<{\centering}}
		\toprule[2pt]
		\multirow{2}{*}{\bf{Methods}}&\multicolumn{4}{c|}{\bf{Combination}}& \multicolumn{5}{C{7.5cm}}{\bf{Dice$^\downarrow$(\%)}}\\
		\cline{2-10}
		&DRPL&CPC& LC & DAE  & AA & LABC & LVBC &LVM &Mean \\ \hline
		Source Only& & & & & 74.51 	&67.57 	&37.30 	&62.97 	&60.59 \\
	     DRPL &\checkmark& & & & 44.56 	&39.05 	&15.75 	&46.12 	&36.37 \\
	     DRPL+CPC&\checkmark& \checkmark& & & 27.15 	&30.08 	&13.41 	&32.26 	&25.73  \\
  	    DRPL+CPC+LC&\checkmark& \checkmark&\checkmark & & 14.90 &11.53 &5.66 	&19.83 	&12.98  \\
	    Ours&\checkmark& \checkmark&\checkmark &\checkmark & \textbf{8.70} 	&\textbf{6.68} 	&\textbf{3.50} 	&\textbf{15.07} 	&\textbf{8.49} \\
	     \bottomrule[2pt]
	     \multicolumn{10}{l}{\footnotesize{\textbf{Source Only}: Baseline model trained with only source domain data. \textbf{DRPL}: Disentangled representation learning.}} \\
	     \multicolumn{10}{l}{\footnotesize{\textbf{CPC}: Content-pattern consistency loss. \textbf{LC}: Label consistency loss.  \textbf{DAE}: Domain-aware pattern encoder. }} \\
	
    	\end{tabular}}
    \end{table}

\section{Discussion}
In this section, we summarize existing SOTA UDA methods and discuss the differences of our proposed framework with them. The core idea of AdaptSegNet \cite{tsai2018learning}, BDL \cite{li2019bidirectional}, and CLAN \cite{luo2019taking} is to align the source domain with the target domain in feature space. These approaches work effectively on data of similar patterns and contents (\emph{e.g.}, natural images), since it is easy to align features from domains closer to each other. However, for our multi-modal medical image data, the difference between modalities presents a sharp change. This sharp change makes the source domain far away from the target domain, and makes these feature-aligning methods fail to deliver a decent performance. In contrast, SIFA \cite{chen2020unsupervised} was tailored for medical scenarios and proposed additional alignments in the image and annotation spaces. Nonetheless, it only works effectively in the easier UDA direction (\emph{i.e.}, MRI to CT on the MMWHS dataset) but yields restricted performance in the reverse, more difficult direction (\emph{i.e.}, CT to MRI on the MMWHS dataset), as shown in Fig.~\ref{fig:Histogram}. 

Making the model insensitive to different patterns for better UDA is a more elegant way. Specifically, we can explicitly make the model aware of the content and pattern of a given image, and then apply the pattern from the target domain to the content from the source domain for training the UDA model. For example, ACE \cite{wu2019ace} utilizes a VGG net to extract the pattern code from the target-domain image and then integrates it into the source-domain image for UDA training. In contrast, DISE \cite{chang2019all} leverages DRPL to extract the content and pattern codes from input of two domains, and adopts the decompose-recompose training strategy for these codes to realize UDA. This method has been verified effective on the BiUDA problem thus has a great potential in medical imaging UDA application. Different from DISE with separated pattern encoders, we propose a unified encoder that helps the model better understand the pattern difference. As shown in Table~\ref{table:abstudy}, with the unified pattern encoder, the framework achieves about 4\% reduction in Dice drop. In addition, under the supervision of the CPC and LC losses, the pattern codes extracted by our method are more effective than those extracted from the VGG net. For this reason, the proposed method outperforms DISE and ACE, as shown in Table~\ref{table:quanti_metric} and Table~\ref{table:quanti_metric1}.

\section{Conclusion}
This work presented a novel, DRPL-based BiUDA framework to address the domain drop problem.
The domain-aware pattern encoder was proposed for obtaining representative pattern codes from both the source and target domains and meanwhile simplifying the network complexity.
To minimize potential data distortion in the process of domain adaptation, the content-pattern consistency loss was devised.
In addition, the label consistency loss was proposed to achieve higher UDA segmentation performance.
The comparative experiments indicates that our framework achieves new SOTA results for the challenging BiUDA segmentation tasks on both MMWHS and MMAS datasets.

%
\bibliographystyle{splncs04}
\bibliography{BIUDA}

\end{document}